\documentclass[10pt,twocolumn,letterpaper]{article}

\usepackage{wacv}
\usepackage{times}
\usepackage{epsfig}
\usepackage{graphicx}
\usepackage{amsmath}
\usepackage{amssymb}
\usepackage{multirow}
\usepackage{booktabs}
\usepackage{multicol}
\usepackage{pifont}
\usepackage{makecell}

\usepackage{algorithm}
\usepackage{algorithmic}
\usepackage{bm}

\def\wacvPaperID{1346} 

\wacvfinalcopy 

\ifwacvfinal
\def\assignedStartPage{9876} 
\fi

\ifwacvfinal
\usepackage[breaklinks=true,bookmarks=false]{hyperref}
\else
\usepackage[pagebackref=true,breaklinks=true,colorlinks,bookmarks=false]{hyperref}
\fi

\ifwacvfinal
\else
\fi

\begin{document}

\title{Deformable Gabor Feature Networks for Biomedical Image Classification}

\author{
Xuan Gong$^{1}$\footnotemark[2], Xin Xia$^{2}$\footnotemark[2], Wentao Zhu$^{3}$, Baochang Zhang$^{2}$, David Doermann$^{1}$, Li'an Zhuo$^{2}$\\
$^1$University at Buffalo $\; ^2$Beihang University $\; ^3$Kwai Inc. \\
{\tt\small \{xuangong, doermann\}@buffalo.edu} \qquad
{\tt\small \{xiaxin, bczhang, lianzhuo\}@buaa.edu.cn} \\
{\tt\small wentaozhu91@gmail.com}
}

\maketitle

\begin{abstract}
In recent years, deep learning has dominated progress in the field of medical image analysis. We find however, that the ability of current deep learning approaches to represent the  complex geometric structures of many medical images is insufficient. One limitation is that deep learning models  require a tremendous amount of data, and it is very difficult to obtain a sufficient amount with the necessary detail. A second limitation is that there are underlying features of these medical images that are well established, but the  black-box nature of existing convolutional neural networks (CNNs) do not allow us to exploit them. In this paper, we revisit Gabor filters and introduce a deformable Gabor convolution (DGConv) to expand deep networks interpretability and enable complex spatial variations. The features are learned at deformable sampling locations with adaptive Gabor convolutions to improve representativeness and robustness to complex objects. The DGConv replaces standard convolutional layers and is easily trained end-to-end, resulting in deformable Gabor feature network (DGFN) with few additional parameters and minimal additional training cost. We introduce DGFN for addressing deep multi-instance multi-label classification on the INbreast dataset for mammograms and on the ChestX-ray14 dataset for pulmonary x-ray images.
\end{abstract}

\renewcommand{\thefootnote}{\fnsymbol{footnote}}
\footnotetext[2]{Equal contribution.}

\section{Introduction}
Automated medical imaging techniques for cancer screening are widely used for lesion analysis~\cite{giger2013breast}, but the traditional pipeline for computer aided diagnosis is typically built based on hand-crafted features~\cite{varela2006use}. These features are not flexible and have poor generalization on unseen data. Deep features, however, are data-driven and are becoming the approach of choice in medical image analysis. Deep learning has achieved great success on skin cancer diagnostics~\cite{esteva2017dermatologist}, organs at risk delineation for radiotherapy~\cite{zhu2018anatomynet} and pneumonia detection from chest x-ray images~\cite{rajpurkar2017chexnet} for example.

One challenge for deep learning is that it is data hungry and often requires expensive and detailed annotation \cite{goodfellow2016deep,song2020learning}. For cancer screening training and validation data in medical images, image-level description of the clinical diagnosis may not be sufficient to train for clinical diagnosis~\cite{10.1007/978-3-030-00934-2_90}. Another challenge arises from CNN itself. CNNs are widely considered black boxes and difficult to interpret. This becomes a greater challenge for weekly supervised learning in biomedical image analysis, whose performance depends highly on powerful representations to handle complicated spatial variations, such as lesion sizes, shapes and viewpoints.   

\begin{figure*}[t]
\centering
\includegraphics[scale=0.40]{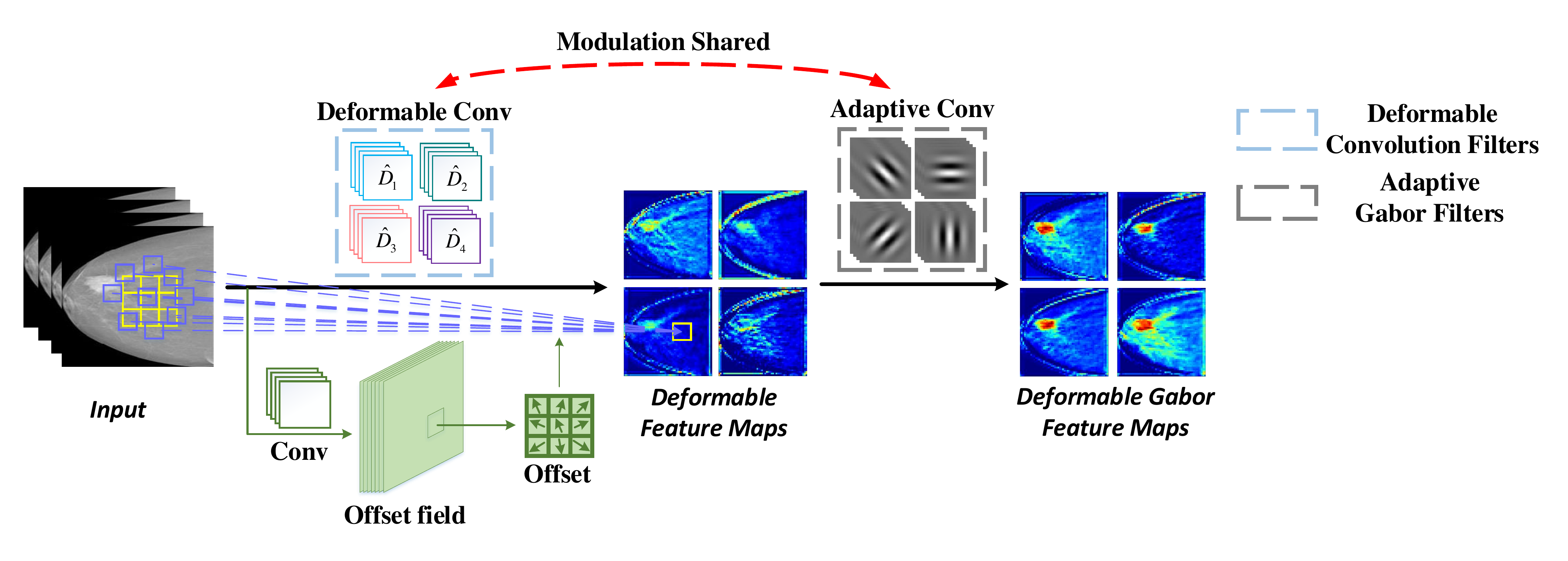}
\caption{The framework of our deformable Gabor Convolution (DGConv).
} 
\label{main}
\end{figure*}

Gabor wavelets~\cite{gabor1946} are widely considered the state-of-the-art hand-crafted feature extraction method, enhancing the robustness of the representation to scale and orientation changes in images. The advantage of Gabor transforms for specific frequency analysis makes them suitable to interpret and resist to dense spatial variations widely existing in biomedical images. Recently, Gabor convolutional networks (GCNs) ~\cite{luan2018gabor} have used Gabor filters to modulate convolutional filters and enhance representation ability of CNNs. \cite{luan2018gabor} only consider rigid transformations of kernels, however, and  not deformable transformations on features that are required for medical image analysis. Thus the robustness of Gabor filters to spatial variations has not been fully investigated to facilitate feature extraction in CNNs. 

On the other hand, deformable convolutional networks (DCNs)~\cite{dai17dcn} augment spatial sampling locations and provide generalized transformations such as anisotropic aspect ratios, demonstrating effectiveness on sophisticated vision tasks such as object detection. We will show that the tailored combination of Gabor filters and deformable convolutions in a dedicated architecture can better characterize the spatial variations and enhance feature representations to facilitate medical image analysis. 

 
In this paper, we investigate deeply into Gabor wavelets with deformable transforms to enhance the networks interpretability and robustness to complex data variations. Unlike previous hand-crafted filters, the newly designed module learns Gabor filters end-to-end, thus improving its adaptiveness to the input data. As illustrated in Figure ~\ref{main}, our deformable Gabor convolution (DGConv) includes deformable convolutions and adaptive Gabor convolutions that share the same modulation information. The deformable convolutions are endowed with local offset transforms to make the feature sampling locations learnable. The adaptive Gabor convolutions further facilitate the capture of visual properties such as spatial localization and orientation selectivity of the input objects, enhancing the generated deformable Gabor features with various dense transformations. To balance the performance and model complexity, we only employ deformable Gabor convolution (DGConv) to extract high level deep features. We integrate this new Gabor module into deep multi-instance multi-label networks, leading to deformable Gabor feature networks (DGFNs) to deal with large variations of objects in medical images. 
The contributions of this work are summarized as follows:
\begin{itemize}
\item Deformable Gabor feature network (DGFN) exploits deformable features and learnable Gabor features in one block to improve the interpretability of CNNs. The noise-resistant property inherited from Gabor features is successfully validated on CIFAR-10 with a $2$\% accuracy improvement over the baseline method. 

\item  DGFN features both the adaptiveness to deformation and robustness to generalize spatial variations common in natural images. Their enhanced representative ability are shown to be beneficial for medical image analysis. 


\item The proposed Gabor module is generic and flexible, which can be easily applied to existing CNNs, such as ResNet and DenseNet. 
\end{itemize}

\section{Related Work}

\subsection{Deformable Convolutional Networks}

CNNs have achieved great success for visual recognition but are inherently limited to spatial variations in object size, pose and viewpoint \cite{lin2017feature,wu2020forest}. One method that has been used to address this problem is data augmentation which adds training samples with extensive spatial variations using random transformations. Robust features can be learned from the data but at the cost of an increased number of model parameters and additional training resources. Another method is to extract spatial invariant features with learned transformations. Ilse \etal~\cite{max2015stn} first proposed spatial transformer networks to learn invariance to translation, scale, rotation and generic warping, giving neural networks the ability to actively and spatially transform feature maps. Deformable convolutional networks (DCNs)~\cite{dai17dcn} introduced offset learning to sample the feature map in a local and efficient manner which can be trained end-to-end. 

\subsection{Gabor Convolution Networks} 
Gabor wavelets~\cite{gabor1946} exhibit strong characteristics of spatial locality, scale and orientation selectivity, and insensitivity to illumination change. The recent rise of deep learning has lead to the combination of Gabor filters and convolution neural networks. Previously Gabor wavelets were only used to initialize  deep networks or  used in the pre-processing~\cite{bodganz,zhong2015high}. \cite{sar2017gfcnn} replaced selected weight kernels of CNNs with Gabor filters to reduce training cost and time. Recent work has integrated Gabor filters into CNNs intrinsically to enhance the resistance of deep learned features to spatial changes~\cite{luan2018gabor}. However, the receptive field of the integrated Gabor filters is fixed and known, and such prior knowledge characterizes limited spatial transformations thus impeding the generalization of complicated spatial variations and new unknown tasks. In this work, we go further by tailoring Gabor filters with learnable modulation masks and deformable transforms. The steerable property of Gabor filters is therefor inherited into the deformable convolutions and its representativeness to spatial variations is fully exploited.

\subsection{Multi-Instance Learning for Weakly Supervised Image Analysis}

There have been a number of  previous attempts to utilize weakly supervised labels to train models for image analysis \cite{song2020unsupervised}.  Papandreou \etal~\cite{papandreou2015weakly} proposed an iterative approach to predict pixel-wise labels in segmentation using image-level labels. Different pooling strategies were proposed for weakly supervised localization and segmentation respectively~\cite{wang2017chestx,bilen2016weakly}. Wu \etal~\cite{wu2015deep} combined CNN with multi-instance learning (MIL) for image auto-annotation. Deep MIL with several efficient inference schemes was proposed for lesion localization and mammogram classification~\cite{zhu2017deep}. Attention based MIL further employed neural attention mechanisms as the inference~\cite{ilse2018attention}. Wan \etal~\cite{Fwan2019MELM} proposed a min-entropy latent model for weakly supervised object detection, which reduces the variance of positive instances and alleviates the ambiguity of the detectors. Unlike previous methods, our method uses a novel feature representation network  to handle large variations of objects in medical images and improve overall image classification. 

\section{Deformable Gabor Convolution}
Without loss of generality, the convolution operation described here is in 2D.

\begin{figure*}[!t]
\centering
\scalebox{1.3}{
\includegraphics[scale=0.37]{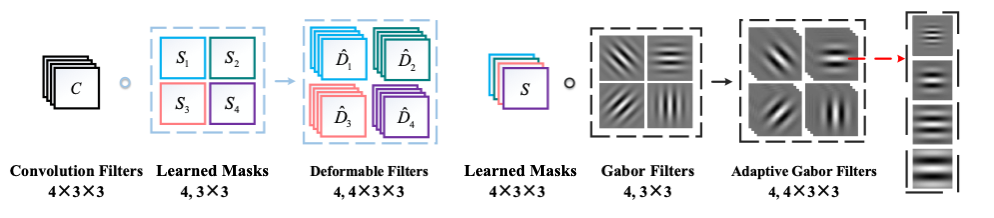}
}
\caption{The modulation process of deformable filters and adaptive Gabor filters. The left shows how convolution filters are modulated by learned masks to generate deformable filters. The right illustrates the generation of adaptive Gabor filters. For illustration convenience, we set the number of learned masks as $V$=4 and the orientation channel of convolution filters and Gabor filters as $U$=4.
}
\label{modulate}
\end{figure*}
\subsection{Deformable and Adaptive Gabor Convolution}
To extract highly representative features, we combine the deformable convolution (DConv) with an adaptive Gabor convolution (GConv) by sharing modulation information. As illustrated in Figure~\ref{modulate}, both the deformable convolution and Gabor transforms are adjusted with the learned masks.

\textbf{Deformable  Convolution:} We are given $U$ standard  convolution filters of size \(H \times H\), which after being modulated by \(V\)  scale kernels of size \(H \times H\), result in   \(U \times V\) modulated convolution filters of size \(H \times H\). We define: 
\begin{equation}
\widehat{\bm{D}}_{u,v} = \bm{C}_{u} \circ \bm{S}_{v},
\end{equation} 
where \(\widehat{\bm{D}}_{u,v} \) indicates the deformable convolution filter, \(\circ \) is element wise product operation, \(\bm{C}_{u} \) is the \(u^{th}\) convolution filter, and \(\bm{S}_{v} \) is the \(v^{th}\) kernel to modulate the convolution filter. In our implementation, the deformable transforms~\cite{dai17dcn}
augment \(\widehat{\bm{D}}_{u,v}\) with translated offsets which are learned from the preceding feature maps through additional convolutions. 

Consider a $3 \times 3$ kernel convolution, \(\mathcal{R} = \{(-1,-1), \cdots,\) \((1,0),(1,1)\}\), with a dilation of 1, for example. Given \({\bm{r}}_{0}\) as the 2D position of output feature and  \(\bm{r}_{n}\) as the location of \(\mathcal{R}\), the deformable convolution filter \(\widehat{\bm{D}}\) can be operated on as follows\footnote{The subscript is omitted for easy presentation.}:
\begin{equation}
{\bm{F_y}(\bm{r}_{0})} = \sum\limits_{\bm{r}_{n} \in \mathcal{R}}{ \widehat{\bm{D}}(\bm{r}_{n}) \times {\bm{F_x}(\bm{r}_{0}+\bm{r}_{n}+\Delta{\bm{r}_{n}})}}  
\label{dconv}
\end{equation}
where \(\bm{F_x}\) and \(\bm{F_y}\) indicate the input and output feature respectively. The learned offset \(\Delta{\bm{r}_{n}}\) updates the offset location to \(\bm{r}_{n}+\Delta{\bm{r}_{n}}\) and adjusts the receptive field of input \(\bm{F_x}\) on which \(\widehat{\bm{D}}\) is implemented.

\textbf{Adaptive Gabor Convolution:} Adaptive Gabor filters are generated from $U$ Gabor filters of size \(H \times H\) with \(V\) learned kernels of size \(H \times H\), where \(U\) indicates the number of orientations of Gabor filters. 
We have:
\begin{equation}
\bm{\widehat{G}}_{v,u} = \bm{S}_{v} \circ \bm{G}_{u},
\end{equation}
where \(\bm{G}_{u}\) is the Gabor filter with orientation \(u\), and \(\bm{\widehat G}_{v,u}  \) is the adaptive Gabor filter corresponding to the \(u^{th}\) orientation and the \(v^{th}\) scale. 
For DGConvs, different layers share the same Gabor filters \(\bm{G} = (\bm{G}_{1}, \cdots, \bm{G}_{U}) \) with various orientations but are adjusted with different information from the corresponding deformable convolution features. 

If the dimensions of the weights in traditional convolution are \(M_0\times N_0\times H \times H\), the dimensions of the learned convolution filters are \(M\times N\times U\times H \times H\) in DGConv,  where \(U\) represents the number of additional orientation channels, \(N\)(\(N_0\)) and \(M\)(\(M_0\))  represent the channel number of the input and output respectively. In DGConv we set \(N=N_0/\sqrt{U}\) and \(M=M_0/\sqrt{U}\) to maintain similar amount of parameters with traditional convolution. Additional parameters in DGConv include \(V\times H \times H\) parameters of mask and \( (2\times H \times H) \times N \times U \times H \times H\) parameters for offset learning, where \(2\times H \times H\) is the channel of offset fields and means that each position of input feature corresponds to an offset of size \(2\times H \times H\) for deformable convolution. 

In DGConv, the number of orientation channels in the input and output feature needs to be \(U\). So the number of orientation channels in the first input feature must be extended to \(U\). For example, if the dimension of original input feature is \(1 \times N \times W \times W\) where \(W \times W\) is the size of input feature, it will be \(U \times N \times W \times W\) after duplicating and concatenating. Thus the new module is light weight and can easily be implemented with a small number of additional parameters. 

\subsection{Forward Propagation}

We use deformable Gabor convolutions (DGConvs) to produce deformable Gabor features. Given the input features \(\bm{F}\), the output Gabor features \(\bm{\widehat F}\) are denoted:
\begin{equation}
\bm{\widehat F} = \mathrm{DGConv}(\bm{F}, \bm{\widehat D}, \bm{\widehat G}),\
\end{equation}
where \(\mathrm{DGConv}\) is the operation which includes deformable convolution filters \(\bm{\widehat D}\) and adaptive Gabor filters \(\bm{\widehat G}\). So the deformable features  \(\bm{E}_v^{(m)}\) and the deformable Gabor features \(\bm{\widehat{F}}_u^{(m)} \) are obtained by:
\begin{equation}
\begin{aligned}
\bm{E}_v^{(m)} = \sum\limits_{n,u} {\bm{F}_u^{(n)} \odot \widehat{\bm{D}}_{u,v}^{(n,m)} }, \quad
\bm{\widehat{F}}_u^{(m)} = \sum\limits_{v} {\bm{E}_v^{(m)} \otimes \widehat{\bm{G}}_{v,u}} ,
\end{aligned}
\end{equation}

where $ \otimes $ denotes the traditional convolution, $\odot$ denotes the deformable convolution shown in Eq. (\ref{dconv}), and  \(n\) and \(m\) denote the number of channels in the input and output features respectively. \(\bm{E}_v^{(m)} \) represents the deformable feature with \(v^{th}\)  modulation in the \(m^{th}\) channel. \(u\) indicates \(\bm{\widehat{F}}_u^{(m)} \) being the \(u^{th}\) orientation response of the deformable Gabor features \(\bm{\widehat F}^{(m)} \). 
Figure~\ref{main} shows that deformable Gabor feature maps reveal better spatial detection results of lesions after the adaptive Gabor convolutions. 

\subsection{Backward Propagation}
During the back propagation in the DGConv, we need to update the kernels \(\bm{C}\) and  \(\bm{S}\), which can be jointly learned. 
The loss function of the network \(\mathcal{L} \) is differentiable within a neighborhood of a point, which will be described  in the next section. We design a novel back propagation (BP) scheme to update parameters:
\begin{equation}
\bm{\delta}_{\bm{S}}  = \frac{{\partial \mathcal{L}}}{{\partial {\bm S}}} = {\frac{\partial \mathcal{L}}{\partial {\widehat {\bm G}}}} \circ \sum\limits_{u = 1}^{U} \bm{G}_{u}, \quad
{\bm S} \leftarrow {\bm S} - \eta_{1} \bm{\delta}_{\bm S},
\end{equation}
where \(\bm{G}_u\) is the Gabor filter with orientation \(u\) and \(\eta_{1} \) denotes the learning rate for $S$. We then fix $S$ and update parameters ${\bm D}$ of deformable convolution filters:
\begin{equation}
\bm{\delta}_{\bm C}  = \frac{{\partial \mathcal{L}}}{{\partial {\bm C}}} = {\frac{{\partial \mathcal{L}}}{{\partial {\bm{\widehat {D}} }}} } \circ \sum\limits_{v = 1}^{V} \bm{S}_{v}, \quad
{\bm D} \leftarrow {\bm C} - \eta_{2} \bm{\delta}_{\bm C}, 
\end{equation}
where \(\bm{S}_v \) is the \(v^{th}\) learned kernel and \(\eta_{2} \) denotes the learning rate of convolution parameters.

\section{Biomedical Image Analysis}\label{sec:GFN4BIOIMAGE}

There are many different ways to formulate problems in biomedical image analysis.  Two of the most common are to classify an entire image as either having a particular condition or not (a binary-label task) and to associate the image with several labels (a multi-label task). To test our deformable Gabor feature network (DGFN), we have identified two representative datasets, the INbreast dataset~\cite{moreira2012inbreast} and the ChestX-ray14 dataset~\cite{wang2017chestx}.

\subsection{The INbreast Dataset}
The INbreast Dataset~\cite{moreira2012inbreast}  is a dataset of mammogram images consisting of 410 images from a total of 115 cases, of which 90 cases are from women with both breasts (4 images per case) and 25 cases are from mastectomy patients (2 images per case)~\cite{moreira2012inbreast}. The dataset includes four types of lesions: masses, calcifications, asymmetries, and distortions. We focus on mass malignancy classification from  mammograms. 

For mammogram classification, the equivalent problem is that if there exists a malignant mass, the mammogram \(\bm{I}\) should be classified as positive. Likewise, a negative mammogram \(\bm{I}\) should not have any malignant masses. If we treat each patch \(\bm{Q}_{k}\) of \(\bm{I}\) as an instance, the mammogram classification is a standard multi-instance learning problem. For a negative mammogram, we expect all the malignant probabilities \(p_{k}\) to be close to 0. For a positive mammogram, at least one malignant probability \(p_{k}\) should be close to 1.

\subsection{The ChestX-ray14 Dataset}

As one of the largest publicly available chest x-ray datasets, ChestX-ray14 consists of 112,120 frontal-view x-ray images scanned from 32,717 patients including many patients with advanced lung diseases~\cite{wang2017chestx}. Each image is labeled with one or multiple pathology keywords, such as atelectasis, or cardiomegaly. This dataset consists of complicated diseases which may have interrelations which can be challenging for the classification task. The ChestX-ray14 dataset has fourteen different labels, so the image classification problem is to associate each instance with a subset of those labels.  This is a multi-instance, multi-label classification problem.

\subsection{Our Approach}
We use the proposed Gabor module to extract highly representative features and design a  multi-instance learning method to deal with deformable Gabor features. 
In this section, we describe the structure of  the deformable Gabor feature networks (DGFNs) for these two problems.

\subsubsection{Multi-Instance Learning for Mammograms} 
\label{sec:milmamm}
 After multiple DGConv layers and rectified linear units, we acquire the last deformable Gabor features \(\bm F\) with multiple channels.  \(\bm{F}_{i,j,:}\) is the feature map for patch \(\bm{Q}_{i,j}\) of the input image, where \(i\) and \(j\) denote the spatial index of the row and column respectively, and \(: \) denotes the channel dimension. We employ a logistic regression model with weights shared across all the patches of the output feature map. A sigmoid activation function for nonlinear transformation is then applied along channels for each element of the output feature map \(\bm{F}_{i,j,:}\) and we slide it over all the pixel positions to calculate the malignant probabilities. The malignant probability of pixel \((i,j)\) in feature space is:
\begin{equation}
p_{i,j} = \mathrm{sigmoid}(\bm{w} \cdot \bm{F}_{i,j,:} + {b}),
\end{equation}
where \(\bm{w} \) is the weight in the logistic regression, \(b\) is the bias, and \( \cdot \) is the inner product of the two vectors \(\bm w \) and \(\bm{F}_{i,j,:}\). \(\bm{w} \) and \({b}\) are shared for different pixel positions \((i,j)\). \(\bm{p} = (p_{i,j})\) is flattened into a one-dimensional vector as \(\bm{p} = (p_{1},p_{2},...,p_{K})\) corresponding to flattened patches \((\bm{Q}_{1},\bm{Q}_{2},...,\bm{Q}_{K})\), where \(K\) is the number of patches. 

 Thus, it is natural to use the maximum component of \(\bm{p}\) as the malignant probability of the mammogram \(\bm{I}\):
\begin{equation}
\begin{aligned}
    p(y = 1|\bm{I}) &= \max \{p_{1},p_{2},...,p_{K}\}, \\
    p(y = 0|\bm{I}) &= 1- p(y = 1|\bm{I}).
\end{aligned}
\end{equation}
The cross entropy-based cost function can be defined as:
\begin{equation}\label{mil}
\mathcal{L} =  - \sum\limits_{n = 1}^N {\log (p(y=y_{n}|\bm{I}_{n}))}, \
\end{equation}
where \(N\) is the total number of mammograms, and \(y_{n} \in \{0,1\}\) is the true label of malignancy for mammogram \(\bm{I}_{n}\) in the training. Typically, a mammogram dataset is imbalanced, where the proportion of positive mammograms is much smaller than negative mammograms, about 1/5 for the INbreast dataset. We therefor introduce a weighted loss:
\begin{equation}\label{wmil}
\mathcal{L} =  - \sum\limits_{n = 1}^N {w({y_{n}})\log (p(y=y_{n}|\bm{I}_{n}))}, \
\end{equation}
where $w(c) = \frac{N}{\sum_{n=0}^{N} {\mathbb{I}} (y_n = c) }$ and ${\mathbb{I}}(\cdot)$ is an indicator function for $y_n$ being label $c$.

\subsubsection{Multi-Instance Multi-Label Learning for Chest X-Rays }
 In our DGFNs for Chest X-Rays dataset, we define a fourteen-dimensional label vector \({\bm{y}_{n}} = [y_n^{1}, y_n^{2}, \cdots, y_n^{C}]\) for \(n^{th}\) image $\bm{I}_{n}$, where \(C = 14\) with binary values, representing either the absence \((0)\) or the presence \((1)\) of a pathology. The \(y_n^{c}\) indicates the presence of an associated pathology in the \(n^{th}\) image where $c=\{1, 2, \cdots, C\}$, while a zero vector \([0, 0, \cdots, 0]\) represents the current x-ray image without any pathology. We consider each pathology as an independent multi-instance learning problem, which is the same as the mammogram classification, to solve the weakly supervised multi-label classification problem. We consider each patch as an instance and the problem can be formulated using equation~(\ref{mil}). If there is no explicit priors on these labels, we can derive the loss function as:
\begin{equation}\label{miml}
\mathcal{L} =  - \sum\limits_{n = 1}^N \sum\limits_{c = 1}^C {\log (p(y=y_{n}^{c}|\bm{I}_{n}))}, \
\end{equation}
where \(N\) is the total number of x-ray images on training set. 
\begin{figure}[!t]
\centering
\includegraphics[scale=0.26]{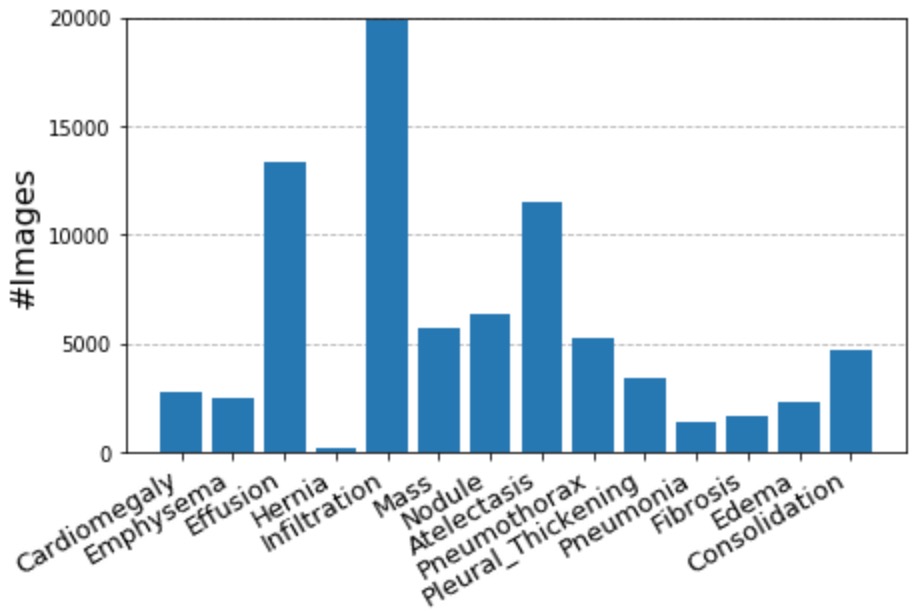}
\caption{Histogram of label frequencies on ChestX-ray14 dataset. The ChestX-ray14 dataset is imbalanced. }
\label{bar}
\end{figure}
As a multi-label problem, we treat all labels equally by defining \(C\) binary cross-entropy loss functions. As the dataset is highly imbalanced as illustrated in Figure~\ref{bar}, we incorporate weights within the loss function based on the label frequency:
\begin{equation}
\mathcal{L} =  -  \sum\limits_{n = 1}^N \sum\limits_{c = 1}^C {w^c(y_n^c) \log (p(y=y_{n}^{c}|\bm{I}_{n}))},
\end{equation}
where $w^c(0) = \frac{N}{\sum_{n=0}^{N} \mathbb{I}(y_n^c = 0) }$ and $w^c(1) = \frac{N}{\sum_{n=0}^{N} \mathbb{I}(y_n^c = 1) }$.

\section{Experiments}
Our deformable Gabor feature networks (DGFNs) are evaluated on the two medical image datasets described above and CIFAR-10 dataset. To balance the performance and training complexity, we use traditional convolution in the first two blocks and  deploy deformable Gabor feature convolution in the following high level features. 

\subsection{Experiments on the INbreast Dataset}\label{sec:INbreast} 
To prepare the data we first remove the background of the mammograms in a pre-processing step using Otsu's segmentation method~\cite{otsu1979threshold}. We then resize the pre-processed mammogram to $224 \times 224$. We use five-fold cross validation with three-fold training, one-fold validation and one-fold testing. We randomly flip the mammograms horizontally, rotate within 90 degree , shift them by 10\% horizontally and vertically, and set a $50 \times 50$ box as 0 for data augmentation. 

The proposed DGFNs employ AlexNet and ResNet18 as the backbones. We use the Adam optimization~\cite{adam2015} algorithm with the initial learning rate of 0.0001 for both \(\eta_{1} \) and \(\eta_{2} \) and weight decay of 0.00005 in the training process. The learning rate decay is set to 10\% for every 100 epochs and the total number of epochs for training is 1000. 

\textbf{Evaluation of $U$ and $V$:} We first perform the experiments on the hyper-parameters $U$ and $V$ to evaluate the additional channel number of orientations and scales. As shown in Table~\ref{V}, given a fixed number of orientations ($U$=4), the average area under the ROC curve (AUC) increases from 79.28\% to 82.53\% when $V$ is changed from $1$ to $5$. Additional evaluation on $U$ shows that DGFN performs better when the number of orientations increases. In the following experiments, we choose $U$=4, $V$=4 to balance the training complexity and performance. 
\begin{table}[]
\centering
\caption{The performance of DGFNs ($U$=4) with different $V$ on INbreast dataset. The last line describes the average training time of one epoch with batch size of 128.
} 
\scalebox{0.97}{
\begin{tabular}{c|c|c|c|c|c}
\Xhline{2\arrayrulewidth}
DGFNs  &$V$=1 &$V$=2 &$V$=3 &$V$=4 &$V$=5\\ \Xhline{2\arrayrulewidth}
AUC (\%) &79.28 &80.72 &81.67 &82.05 &82.53 \\ \hline
Times (s) &2.96 &4.03 &5.87 &6.85 &7.92 \\ 
\Xhline{2\arrayrulewidth}
\end{tabular}}
\label{V}
\end{table}

\begin{figure}[!t]
\centering
\includegraphics[scale=0.52]{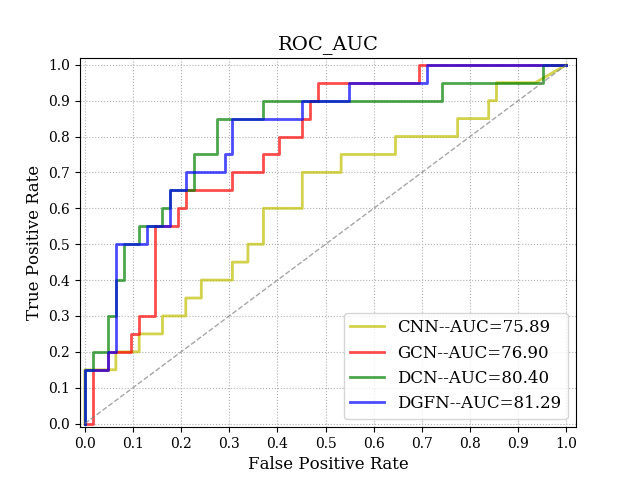}
\caption{AUC comparison on INbreast-deform. All the networks are of similar model sizes with CNN $0.70$M, GCN $0.70$M, DCN $0.83$M and DGFN $0.98$M. }
\label{auc}
\end{figure}

\textbf{Deformation Robustness and Model Compactness:} To validate the networks robustness to deformation, we generate a deformable version of the dataset called INbreast-Deform by sampling 50 images with random scale and rotation for each test sample of the INbreast dataset. Scale factors are in the range \([0.5, 1.5)\), and rotation angles are in the range \([0, 2\pi)\).  The results in Table~\ref{params} confirm that our DGFNs outperform CNNs even with fewer parameters by reducing the channel size of features in the network. When compared to CNNs with a similar number of parameters, DGFNs with kernel stage 8-16-32-64 and 16-32-64-128 obtain larger AUC improvements from 75.89\% to 81.29\% and from 78.26\% to 83.30\% respectively. Figure~\ref{auc} is the comparison of the  average  area  under  the  ROC  curve  (AUC) of CNN, GCN, DCN and DGFN with similar sizes around 0.70-0.98M. DGFNs also achieve better performance than baseline methods including GCNs and DCNs. Thus DGFN enhances the robustness to spatial variations widely existing in biomedical images and largely reduces the complexity and redundancy of the network. %

\begin{table}[]
\footnotesize
\centering
\caption{Comparisons among CNNs, GCNs, DCNs and DGFNs on INbreast-Deform. } 
\scalebox{1.08}{
\begin{tabular}{cccc}
\Xhline{2\arrayrulewidth}
\multirow{2}{*}{Backbone} & \multirow{2}{*}{\begin{tabular}[c]{@{}l@{}}Kernel \\ Stages\end{tabular}} & \multirow{2}{*}{AUC (\%)} &\multirow{2}{*}{\#Params (M)}  \\
                          &                                                                           &                      &  \\ \Xhline{2\arrayrulewidth}
\multirow{2}{*}{ResNet18}   &  \begin{tabular}{@{}c@{}}16-32-64-128\end{tabular}                                                                         & 75.89                    & 0.70\\
                          & \begin{tabular}{@{}c@{}}32-64-128-256\end{tabular}                                                                          & 78.26                     & 2.80 \\ \Xhline{2\arrayrulewidth}
\multirow{2}{*}{\begin{tabular}{@{}c@{}}ResNet18 \\(GCNs)\end{tabular}}   &  \begin{tabular}{@{}c@{}}8-16-32-64\end{tabular}                                                                         & 76.90                     & 0.70 \\
                          & \begin{tabular}{@{}c@{}}16-32-64-128\end{tabular}                                                                          & 79.16                     & 2.80 \\ \hline
\multirow{2}{*}{\begin{tabular}{@{}c@{}}ResNet18 \\(DCNs)\end{tabular}}   &  \begin{tabular}{@{}c@{}}16-32-64-128\end{tabular}                                                                         & 80.40                     & 0.83\\
                          & \begin{tabular}{@{}c@{}}32-64-128-256\end{tabular}                                                                          & 82.03                     & 3.05 \\ \hline                          
\multirow{3}{*}{\begin{tabular}{@{}c@{}}ResNet18 \\(DGFNs)\end{tabular}}    &\begin{tabular}{@{}c@{}}8-16-32-32\end{tabular}                                                                           & 77.59                     & 0.53 \\
                          & \begin{tabular}{@{}c@{}}8-16-32-64\end{tabular}                                                                          & \textbf {81.29}                    & 0.98 \\
                          & \begin{tabular}{@{}c@{}}16-32-64-128\end{tabular}                                                                      &\textbf{83.30}                    &3.40 \\ \Xhline{2\arrayrulewidth}
\end{tabular}}
\label{params}
\end{table}

\begin{figure}[!t]
\centering
\includegraphics[scale=0.56]{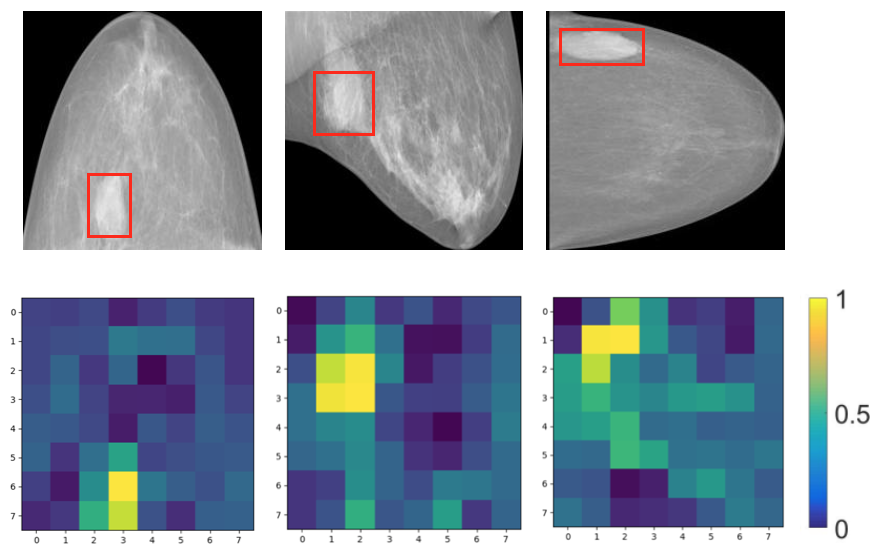}
\caption{Malignant probability of each patch on INbreast dataset. The feature map has $8\times8$ patches. }
\label{heatmap}
\end{figure}

On the INbreast dataset, we combine DGFN with the multi-instance loss explained in section~\ref{sec:milmamm}. As shown in Figure \ref{heatmap}, our designed method can extract features and pinpoint the malignant region effectively. DGFNs with AlexNet and ResNet18 are compared with previous state-of-the-art approaches based on  sparse multi-instance learning (Sparse MIL)~\cite{zhu2017deep}. 
As shown in Table~\ref{tab:inbreast},
DGFNs have enhanced representative ability and achieve better AUC than previous approaches. 

\begin{table}[]
\fontsize{9pt}{11pt}\selectfont\centering
	\caption{Comparisons on INbreast dataset. DGFN with ResNet18 yields the best performance.}
\scalebox{1.00}{
\begin{tabular}{c|c|c}
 \Xhline{2\arrayrulewidth}

Methods      & Acc (\%)         & AUC (\%)             \\  \Xhline{2\arrayrulewidth}
AlexNet+Label Assign. MIL~\cite{zhu2017deep}  &84.16  & 76.90            \\
\hline
AlexNet+ DGFN+ MIL  &86.22  & 78.12            \\
\hline
ResNet18+ DGFN+ MIL  &88.61  & 82.19           \\
\hline
\hline
Pretrained AlexNet+Sparse MIL~\cite{zhu2017deep}  &90.00  & 85.86            \\
\hline
Pretrained AlexNet+ DGFN + MIL  &91.34 &\textbf{87.22}          \\ \hline 
Pretrained ResNet18 + DGFN + MIL  &93.18  &\textbf{88.05}         \\ \Xhline{2\arrayrulewidth}
\end{tabular}}
\label{tab:inbreast}
\end{table}

\begin{figure}[t]
\centering
\includegraphics[scale=0.41]{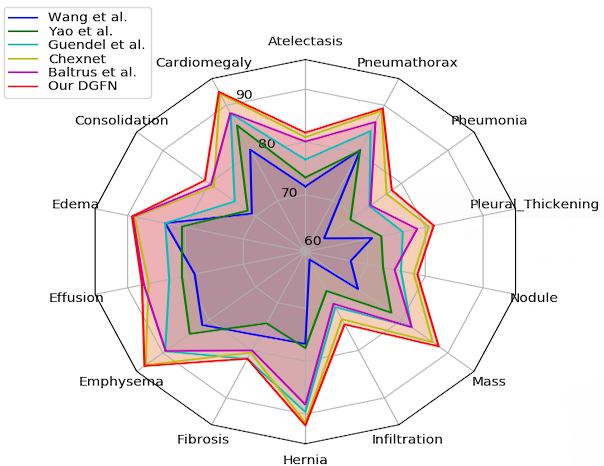}
\caption{AUC (\%) comparisons of our best model with state-of-art methods on ChestX-ray14 dataset.} 
\label{chex_comp}
\end{figure}
\begin{table*}
\fontsize{9pt}{11pt}\selectfont\centering
	\caption{AUC (\%) comparisons of DGFN with Off-The-Shelf (OTS) and Fine-Tune (FT) state-of-art methods on ChestX-ray14 dataset. Bold text emphasizes the highest value among each group.}
\scalebox{1.0}{	
\begin{tabular}{c|c c c|c c c c c}
\Xhline{2\arrayrulewidth}
\multirow{2}{*}{Pathology} & \multicolumn{3}{c|}{Off-The-Shelf} & %
    \multicolumn{5}{c}{Fine-Tune}  \\
\cline{2-9}
&Yao et al.  &Baltruschat et al. &DGFN &Wang et al. &Guendel et al. &Chexnet &Baltruschat et al.  & DGFN \\
&(2017) &(2019) &(Ours) &(2017) &(2018) &(2018) &(2019) &(Ours)\\
\Xhline{2\arrayrulewidth}
Atelectasis &73.3 &73.2 &\textbf{78.04} &71.6 &76.7  &80.94  &80.1 &\textbf{81.78} \\
Cardiomegaly &85.8 &75.9 &\textbf{89.01} &80.7 &88.3  &92.48 &88.4 &\textbf{92.84}  \\
Consolidation &71.7 &75.3 &\textbf{79.09} &70.8 &74.5  &79.01 &79.6 &\textbf{80.91}  \\
Edema &80.6 &85.7 &\textbf{87.21} &83.5 &83.5 &88.78  &89.1 &\textbf{89.25}  \\
Effusion &80.6 &80.6 &\textbf{86.89} &78.4 &82.8  &86.38  &87.2 &\textbf{87.51}  \\
Emphysema &\textbf{84.2} &79.8 &81.96 &81.5 &89.5  &93.71  &89.4 &\textbf{93.97}  \\
Fibrosis &74.3 &73.9 &\textbf{76.08} &76.9 &81.8   &80.47  &80.0 &\textbf{81.75}  \\
Hernia &77.5 &\textbf{81.9} &77.83 &76.7 &89.6 &91.64  &88.2 &\textbf{92.15}  \\
Infiltration &67.5 &67.0 &\textbf{68.49} &60.9 &70.9 &73.45 &70.2 &\textbf{74.52}  \\
Mass &\textbf{77.8} &68.6 &76.32 &70.6 &82.1  &86.76 &82.2 &\textbf{88.03}  \\
Nodule &72.7 &66.5 &\textbf{67.19} &67.1 &75.8  &78.02  &74.7 &\textbf{78.65}  \\
Pleural\(\_\)Thickening &72.4 &70.8 &\textbf{73.32} &70.8 &76.1 &80.62 &78.6 &\textbf{81.47}  \\
Pneumonia &69.0 &68.3 &\textbf{72.83} &63.3 &73.1  &76.80 &73.3 &\textbf{77.91}  \\
Pneumathorax &80.5 &79.1 &\textbf{83.17} &80.6 &84.6 &88.87 &86.5 &\textbf{89.36}  \\ \Xhline{2\arrayrulewidth}
Average &76.1 &74.8 &\textbf{78.39} &73.8 &80.7 &84.17 &82.0 &\textbf{85.01}  \\ \Xhline{2\arrayrulewidth}
\end{tabular}}
\label{tab:chectxray14}
\end{table*}

\begin{table}
\footnotesize
\centering
\caption{Comparisons among CNNs, GCNs, DCNs and DGFNs on CIFAR-10 and CIFAR-10-Noise.}
\scalebox{0.97}{
\begin{tabular}{ccccc}
\Xhline{2\arrayrulewidth}
Methods &\begin{tabular}{@{}c@{}}Kernel  \\ Stages\end{tabular} &Acc (\%) &\begin{tabular}{@{}c@{}}Acc with \\ noise (\%) \end{tabular} &\begin{tabular}{@{}c@{}}\#Params \\ (M)\end{tabular}  \\ \Xhline{2\arrayrulewidth}

\multirow{1}{*}{ResNet18}  & \begin{tabular}{@{}c@{}}32-64-128-256\end{tabular}                                                                          & 90.74                     & 70.72      &2.80\\ \Xhline{2\arrayrulewidth}
\multirow{2}{*}{\begin{tabular}{@{}c@{}}ResNet18 \\(GCNs)\end{tabular}}    &\begin{tabular}{@{}c@{}}8-16-32-64\end{tabular}                                                                           & 88.3                     & 72.81                &0.70\\
                          & \begin{tabular}{@{}c@{}}16-32-64-128\end{tabular}                                                                          &  89.37                    & 74.69       &2.80\\ \hline
\multirow{2}{*}{\begin{tabular}{@{}c@{}}ResNet18 \\(DCNs)\end{tabular}}    &\begin{tabular}{@{}c@{}}16-32-64-128\end{tabular}                                                                           & 88.92                     & 74.30              &0.83\\
                          & \begin{tabular}{@{}c@{}}32-64-128-256\end{tabular}                                                                          &  89.79                    & 78.96      &3.05\\ \hline
\multirow{2}{*}{\begin{tabular}{@{}c@{}}ResNet18 \\(DGFNs)\end{tabular}}    &\begin{tabular}{@{}c@{}}8-16-32-64\end{tabular}                                                                           & 89.59                     & \textbf{76.75}                 &0.98\\
                          & \begin{tabular}{@{}c@{}}16-32-64-128\end{tabular}                                                                          &  \textbf{91.03}                    & \textbf{80.12}       &3.40\\  \Xhline{2\arrayrulewidth}
\end{tabular}}
\label{tab:cifar10}
\end{table}

\begin{figure}[!t]
\centering
\includegraphics[scale=0.42]{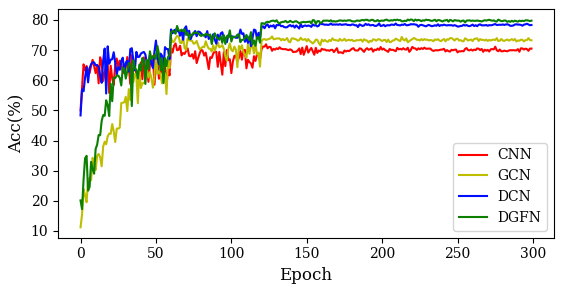}
\caption{Comparisons of accuracy on CIFAR-10-Noise. Note that the four models are of similar size with CNN $2.80$M, GCN $2.80$M, DCN $3.05$M and DGFN $3.40$M.  }
\label{noiseacc}
\end{figure}

\subsection{Experiments on the ChestX-ray14 Dataset}\label{sec:ChestX} 

 We resize the x-ray images from \(1024 \times 1024\) to \(224 \times 224\) to reduce the computational cost and normalize them based on the mean and standard deviation of images from the ImageNet training set~\cite{deng2009imagenet}. In our experiments, we employ a DenseNet121~\cite{huang2017densely} as the backbone of our DGFN on ChestX-ray14 dataset. We resize the images to \(224 \times 224\) and further augment the training data with random rotation and horizontal flipping. During training we use stochastic gradient descent (SGD) with momentum 0.9 and batch size 16. We use initial learning rates of 0.001 that are decayed by a factor of 10 each time when the validation loss has no improvement.

We used the official split released by Wang \etal ~\cite{wang2017chestx} with 70\% training, 20\% testing and 10\% validation. While Yao \etal ~\cite{Yao2017LearningTD} and Chexnet ~\cite{rajpurkar2017chexnet} randomly split the dataset and ensure that there is no patient overlap between the splits. Yao \etal ~\cite{Yao2017LearningTD} noted that there is insignificant performance difference with different random splits. Thus it is a fair comparsion. We divide the compared methods into Fine-Tune (FT) and Off-The-Shelf (OTS) based on whether it used additional data for training. Guendel \etal~\cite{guendel2018learning} used another fully annotated dataset-PLCO Dataset~\cite{plco} to facilitate training. While our DGFN and other comparable fine-tuned methods~\cite{rajpurkar2017chexnet,wang2017chestx,Bal18chesx} are initialized with ImageNet. Table~\ref{tab:chectxray14} demonstrates that among the group labeled fine-tune, DGFN with DenseNet121 outperforms \cite{rajpurkar2017chexnet,wang2017chestx,Bal18chesx} on all fourteen pathologies from the ChestX-ray14 dataset. 
Among the group labeled off-the-shelf, DGFN achieves average AUC of 78.39\% and performs better on 11 out of 14 pathologies than other methods ~\cite{Yao2017LearningTD,Bal18chesx}. Figure~\ref{chex_comp} illustrative effectiveness of DGFN to enhance variant representations, which is potentially of great help on automated biomedical image analysis.

\subsection{Experiments on the CIFAR-10 Dataset} 
To verify the effectiveness of DGFN on the natural image dataset, we conduct extensive experiments on CIFAR-10 as well as CIFAR-10 with noise. We generate a noisy version of CIFAR-10 called CIFAR-10-Noise by replacing the pixel value with 255 at a probability of 1\% percentage to test the network's robustness to random Gaussian noise. We train on CIFAR-10 with random flipping and crop as augmentation . We test on CIFAR-10 and CIFAR-10-Noise respectively. We use ResNet18 as the backbone and use SGD optimization with the initial learning rates as 0.05. The batch size is set as 128 and the total number of training epochs is 300. Figure~\ref{noiseacc} is the comparison of test accuracy on CIFAR10-Noise with CNN, GCN, DCN and DGFN of similar sizes. Table~\ref{tab:cifar10} shows that the proposed DGFNs outperform the baseline on CIFAR-10-Noise. With a similar number of parameters, DGFN with kernel stage 16-32-64-128 achieves a 2\% accuracy improvement beyond DCN, demonstrating its own superior robustness to random Gaussian noise common on natural images.

\section{Conclusion}
We have presented a deformable Gabor feature network (DGFN) to improve the robustness and interpretability for weakly supervised biomedical image classification. DGFN integrates adaptive Gabor filters into deformable convolutions, thus sufficiently characterizes spatial variations in objects and extracts discriminative features for various categories. Experiments show the DGFN is resistant to Gaussian noise and the architecture is both efficient and compact. DGFN is easily integrated into multi-instance, multi-label learning to facilitate the classification of biomedical image with great variations of sizes and shapes of the lesions. Extensive experiments demonstrate the effectiveness of DGFNs on both the INbreast dataset and the ChestX-ray14 dataset. 
\section*{Acknowledgements}
Baochang Zhang is the corresponding author. This study was supported by Grant NO.2019JZZY011101 from the Key Research and Development Program of Shandong Province to Dianmin Sun. 
{\small
\bibliographystyle{ieee_fullname}
\bibliography{gaborbib}
}

\end{document}